\documentclass[sigconf]{acmart}

    \usepackage{tikz}
    \usetikzlibrary{decorations.pathreplacing}
    \usetikzlibrary{shapes,arrows,positioning,fit,chains}
    \usepackage{subcaption}
    \usepackage{soul}

\AtBeginDocument{%
  \providecommand\BibTeX{{%
    \normalfont B\kern-0.5em{\scshape i\kern-0.25em b}\kern-0.8em\TeX}}}
    
\setcopyright{none}
\renewcommand\footnotetextcopyrightpermission[1]{}
\makeatletter
\renewcommand\@formatdoi[1]{\ignorespaces}
\makeatother

\begin{document}

\title{Case-Based Reasoning for Assisting Domain Experts in Processing Fraud Alerts of Black-Box Machine Learning Models}


\author{Hilde J.P. Weerts}
\email{h.j.p.weerts@student.tue.nl}
\affiliation{%
  \institution{Eindhoven University of Technology}
  \city{Eindhoven}
  \country{The Netherlands}
}

\author{Werner van Ipenburg}
\email{werner.van.ipenburg@rabobank.nl}
\affiliation{%
  \institution{Rabobank Nederland}
  \city{Zeist}
  \country{The Netherlands}}

\author{Mykola Pechenizkiy}
\email{m.pechnizkiy@tue.nl}
\affiliation{%
  \institution{Eindhoven University of Technology}
  \city{Eindhoven}
  \country{The Netherlands}
}

\renewcommand{\shortauthors}{Weerts, et al.}

\begin{abstract}
In many contexts, it can be useful for domain experts to understand to what extent predictions made by a machine learning model can be trusted. In particular, estimates of trustworthiness can be useful for fraud analysts who process machine learning-generated alerts of fraudulent transactions. In this work, we present a case-based reasoning (CBR) approach that provides evidence on the trustworthiness of a prediction in the form of a visualization of similar previous instances. Different from previous works, we consider similarity of local post-hoc explanations of predictions and show empirically that our visualization can be useful for processing alerts. Furthermore, our approach is perceived useful and easy to use by fraud analysts at a major Dutch bank.
\end{abstract}




\maketitle

\section{Introduction}
Machine learning models are increasingly applied in real-world contexts, including detection of fraudulent transactions. Often, the best performing models are complex models such as deep neural networks and ensembles. Despite their excellent performance, these models are not infallible, and predictions may require post-processing by human domain experts. However, as the complexity of the model increases, it can become more difficult for human domain experts to assess the correctness of a prediction. In such cases, domain experts who post-process predictions can benefit from evidence on the \textit{trustworthiness} of the model's prediction. Moreover, this type of evidence could be useful to identify when the model is no longer trustworthy due to concept drift~\cite{zliobaite2015appsCD,DBLP:journals/csur/GamaZBPB14}, which is particularly relevant in the context of fraud detection. A straightforward indicator of the trustworthiness of a prediction is the model's own reported confidence score. However, raw confidence scores are often poorly calibrated~\citep{Kuleshov2015}, which means they can be misleading for human domain experts. Furthermore, since the models are not perfect, and in case of heavily imbalanced problems including fraud detection far from being perfect, even calibrated confidence scores can be inaccurate and hence misleading for processing of fraud alerts. Recent explanation methods, including e.g.\ SHAP~\cite{Lundberg2017}, LIME~\cite{Ribeiro2016}, Anchor~\cite{Ribeiro2018} can potentially help domain experts in determining to what extend the model's predictions can be trusted. For example, domain experts can look at local feature importance of the alert that is being processed or at a local surrogate model that mimics the behavior of the global model in the neighborhood of the alert. However, there is lack of empirical evidence that would illustrate the utility of such approaches for alert processing tasks. Our recent user study on the utility of SHAP for processing alerts suggests that SHAP explanations alone do not contribute to better decision-making by domain experts~\cite{Weerts_XAI19}.  

\subsubsection*{Approach} 
In the present paper, we introduce a case-based reasoning (CBR) approach to provide domain experts with evidence on the trustworthiness of a prediction. The proposed approach consists of two steps: (1) retrieve the $k$ most similar instances to the query instance and (2) visualize the similarity as well as the true class of the retrieved neighbors. If the true class of similar instances corresponds to the prediction of the model, this provides evidence on the trustworthiness of the model's prediction, and the other way around. 
An important consideration of any nearest-neighbor type approach is the distance function. A straightforward notion of similarity in our scenario is similarity in feature values. However, instances with very similar feature values may be treated very differently by the model. Thus, it may be more useful for alert processing to consider similarity in \textit{local feature contributions}. That is, we can consider whether the model's predictions of an instance can be explained in a similar way as the prediction corresponding to the alert. Different from previous works, we consider distance functions that take into account similarity in feature values, local explanations, and combinations thereof.

\subsubsection*{Empirical Evaluation}
In simulated user experiments, we empirically show that our approach can be useful for alert processing. In particular, the usage of a distance function that considers similarity in local feature contributions often results in the best performance. Furthermore, a usability test with fraud analysts at a major Dutch bank indicates that our approach is perceived useful and easy to use.

\subsubsection*{Outline}
The present paper is structured as follows. Section~\ref{sec:relwork} covers related work. In Section~\ref{sec:cbr}, we introduce our CBR approach. In Section~\ref{sec:eval}, we present the results of an empirical evaluation of our approach. We discuss concluding remarks in Section~\ref{sec:concl}.

\section{Related Work}
\label{sec:relwork}
The basis of our approach is CBR: similar problems have similar solutions and previous solutions can be used to solve new problems \cite{Kolodner1992}. CBR decision support systems became popular during the nineties and many different case-based explanations (CBE) have been suggested in that context \cite{Srmo2005}. 

Arguably the most straightforward CBE method is to retrieve the most similar cases. For example, \citet{Nugent2005} propose to retrieve the most similar instance to the current case, weighted by the local feature contributions of the query instance. The proposed distance function is intuitive, but its utility is not empirically tested. In the related field of $k$-Nearest Neighbor ($k$-NN) classification, the importance of the distance function has been long recognized. Different weight-setting algorithms have been proposed to improve the performance of $k$-NN algorithms. In particular, different algorithms can be distinguished based on whether weights are applied \textit{globally} (i.e. feature importance across all instances) or \textit{locally} (i.e. feature importance per instance or a subgroup of instances) \citep{Wettschereck1997}. In our experiments, we consider several distance functions, including unweighted, locally weighted, and globally weighted functions.

Similar to our goal, \citet{Jiang2018} propose to capture trustworthiness into a \textit{trust score}: the ratio between the distance from the query instance to the nearest class different from the predicted class and the distance to the predicted class. However, the utility of the trust score for human users is not evaluated. In particular, summarizing trustworthiness in a single score makes it impossible for users to determine whether the derivation of the score aligns with their domain knowledge. Moreover, the score can take any value, which can make it difficult to interpret by novice users. 

\section{Case-Based Reasoning Approach}
\label{sec:cbr}
\begin{figure}[ht]
    \centering
    
    \begin{subfigure}[t]{\linewidth}
        \centering
        \includegraphics[scale=0.15,trim=10 10 10 10,clip]{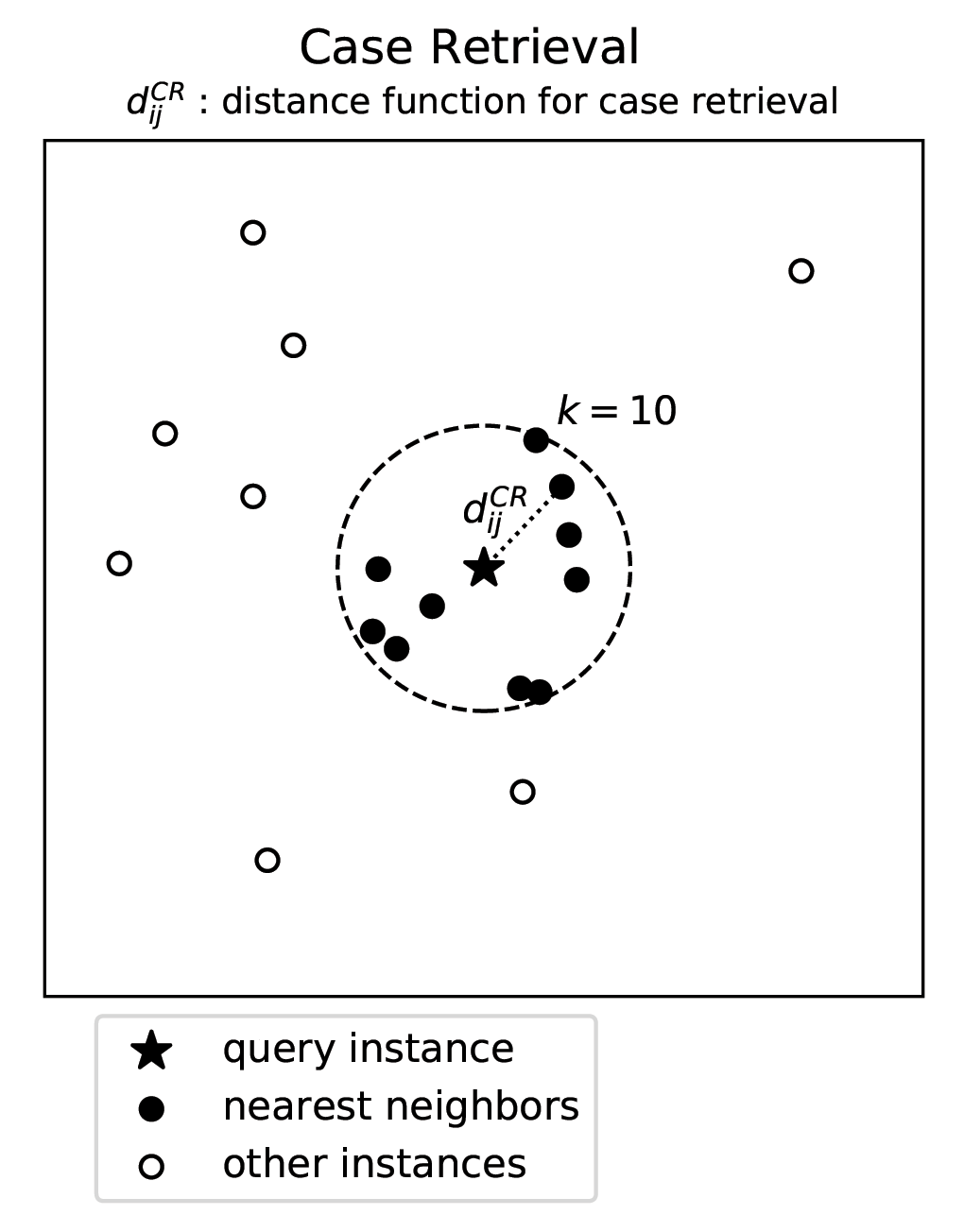}
        \caption{\textit{Case Retrieval}. Retrieve the $k$ instances most similar to the query instance.}
        \label{fig:cbrapproachcr}
    \end{subfigure}
    
    \begin{subfigure}[t]{\linewidth}
        \centering
        \includegraphics[scale=0.15,trim=10 10 10 10,clip]{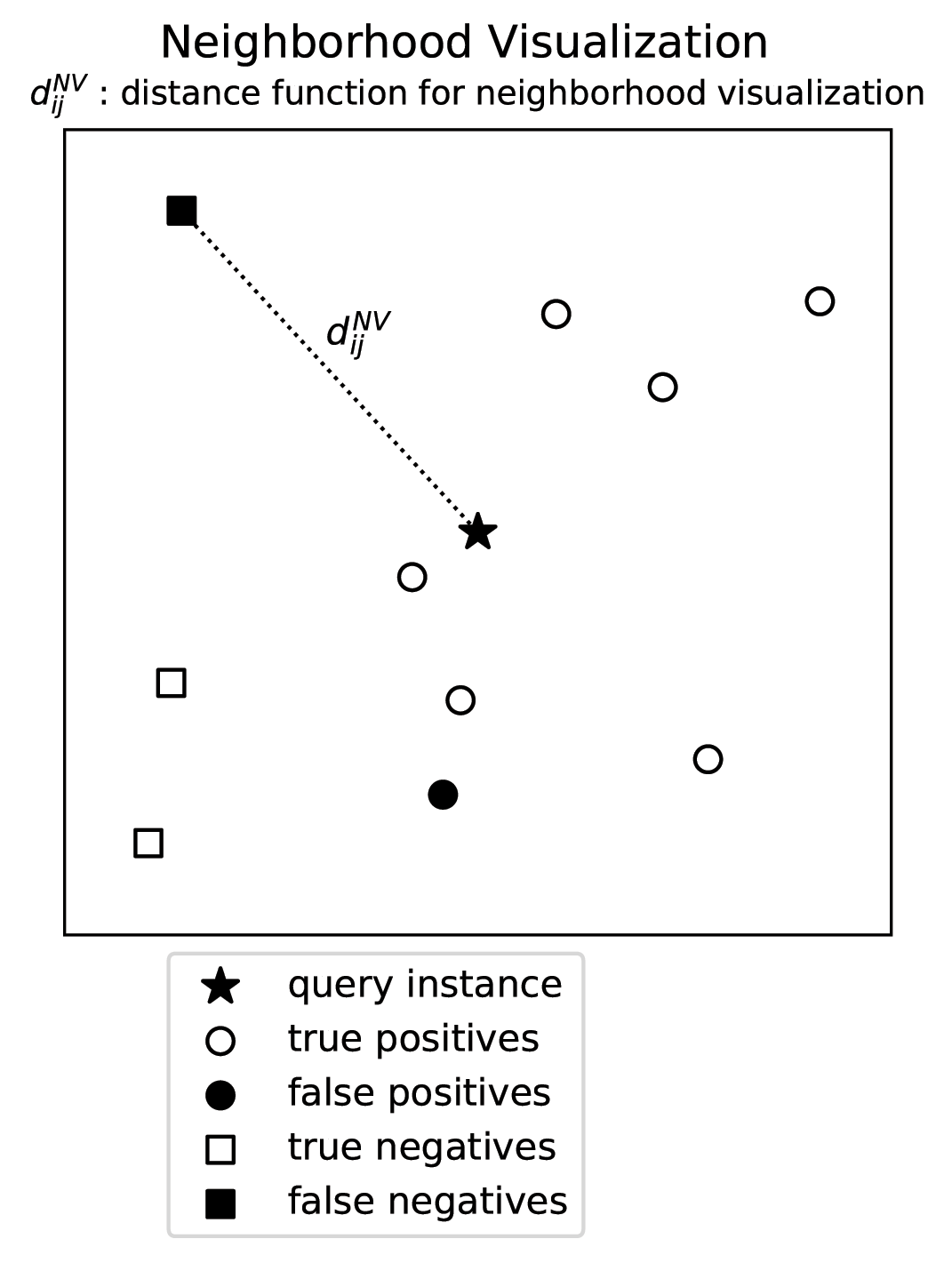}
        \caption{\textit{Neighborhood Visualization}. Visualize how similar the retrieved $k$ neighbors are as well as their true class.} \label{fig:cbrapproachnv}
    \end{subfigure}
    
    \caption{The two stages of the CBR approach for estimating the trustworthiness of a local prediction.}
    \label{fig:cbrapproach}
\end{figure}

For a given instance, which we will refer to as the \textit{query instance}, our approach consists of the following two steps (see Figure~\ref{fig:cbrapproach}):
\begin{enumerate}
    \item \textit{Case retrieval}. Retrieve the $k$ instances from the case base that are most similar to the query instance. The appropriate distance function may be different for different problems. The case base consists of instances for which the true class is known. 
    \item \textit{Neighborhood visualization}. Visualize the $k$ retrieved instances as well as the query instance as points in a scatter plot, such that:
    \begin{enumerate}
        \item the distance between any two instances corresponds to their similarity according to a distance function;
        \item the colors of the neighbors correspond to their true classes.
    \end{enumerate} 
\end{enumerate}
The number of retrieved cases, $k$, is a user-defined parameter; i.e. the user can choose how many instances they would like to retrieve. Note that a different distance function can be used for step (1) and (2). In Section~\ref{sec:evalsim}, we empirically test which combination of distance functions are most useful in each step for several benchmark data sets as well as a real-life fraud-detection data set. 

\subsection{Case Retrieval}
In the case retrieval stage, we retrieve instances from the case base that are similar to the instance we are trying to explain.

The \textit{case base} consists of instances for which the ground truth is known at the time the machine learning model is trained. When the amount of historical data is relatively small, all instances can be added to the case base. Otherwise, sampling or prototyping approaches may be used to decrease the size of the case base.

Our approach assumes that the true class of instances in the case base is known. In some contexts, such as money laundering, the true class is typically not known for all instances. When instances whose true class has not been verified are added to the case base, this should be made explicit to the user in the neighborhood visualization, e.g. by means of a different color.

\subsection{Neighborhood Visualization}
Rather than just returning the retrieved neighbors to the user, we visualize the neighborhood in a two dimensional scatter plot (see Figure~\ref{fig:cbrapproachnv}). The distance between any two instances $i$ and $j$ in the visualization roughly corresponds to the dissimilarity of two instances, given by the used distance function. Similar to the approach taken by \citet{McArdle2003}, we compute the coordinates of the scatter plot using \textit{multidimensional scaling} (MDS) \cite{Kruskal1964}.

In addition, colors or shapes can be used to visualize the model's performance for each of the retrieved instances. The user can use this information to determine whether the prediction of the query instance is trustworthy or not. For example, if many of the most similar neighbors are false positives, this decreases the trustworthiness of an alert.

The visualization can be further extended based on the application context. For example, we know that fraud schemes change over time. Hence, transactions that occurred a long time ago may be less relevant than newer transactions. In this scenario, a valuable extension could be to visualize the maturity of retrieved instances using e.g. an time-based color scale or a time filter. Another interesting extension would be to add \textit{counterfactual instances} to the visualization. Counterfactual instances are perturbed versions of the query instance that received a different prediction from the model \citep{Wachter2017}. Adding such perturbed instances to the visualization may help the domain expert to identify and inspect the decision boundary of the classification model. We leave this to future work.

\subsection{Distance Functions}
\label{sec:dist}
\begin{figure*}[ht]
\centering
  \setbox1=\hbox{\includegraphics[scale=0.09]{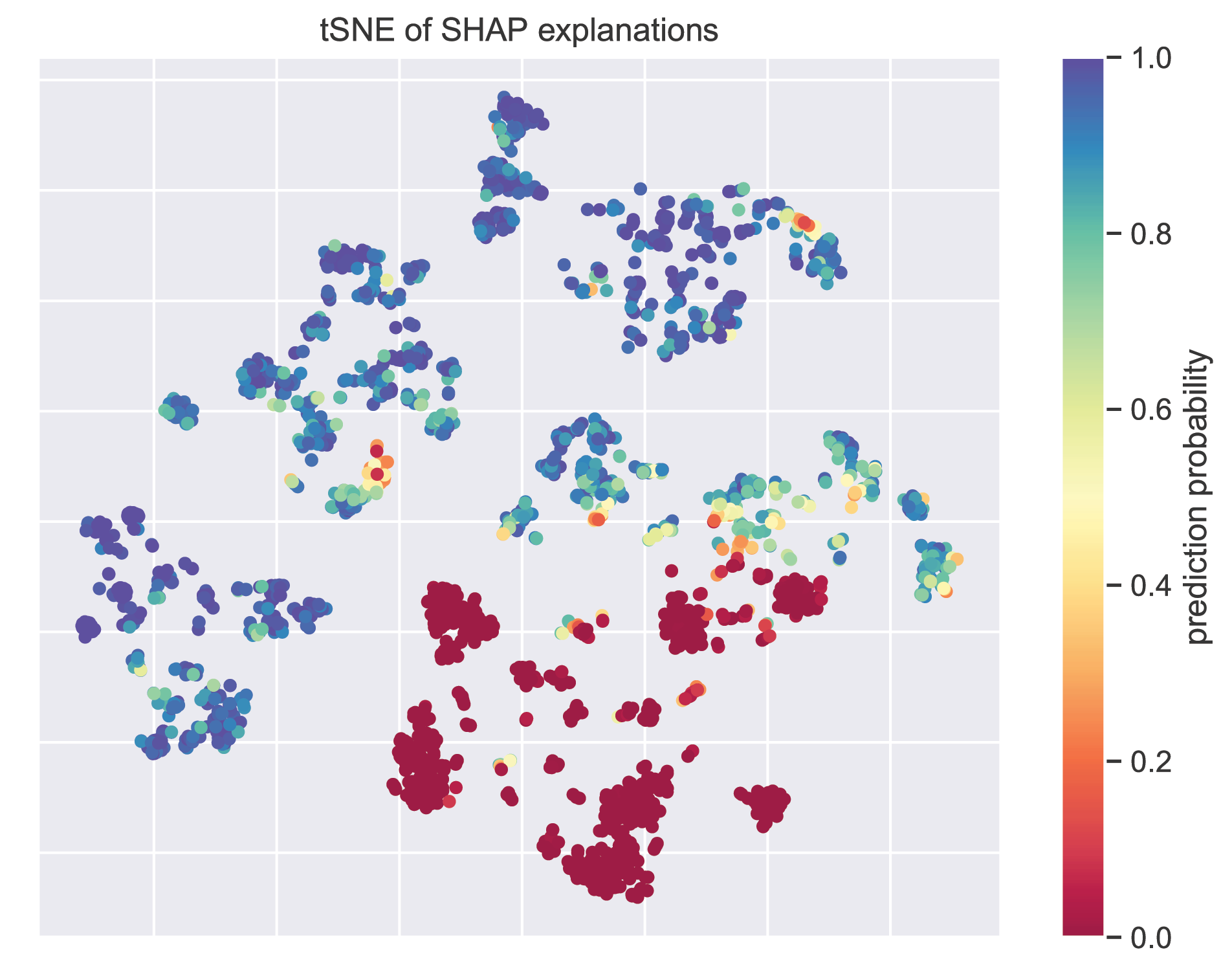}}
  \begin{subfigure}[t]{0.32\textwidth}
    \includegraphics[scale=0.09,trim=10 10 10 10,clip]{figures/tsne/tsne_rabo.png}
    \caption{Model's Confidence}
    \label{fig:tsne_rabo}
  \end{subfigure}
  \begin{subfigure}[t]{0.32\textwidth}
    \centering
    \includegraphics[height=\ht1,trim=10 10 10 10,clip]{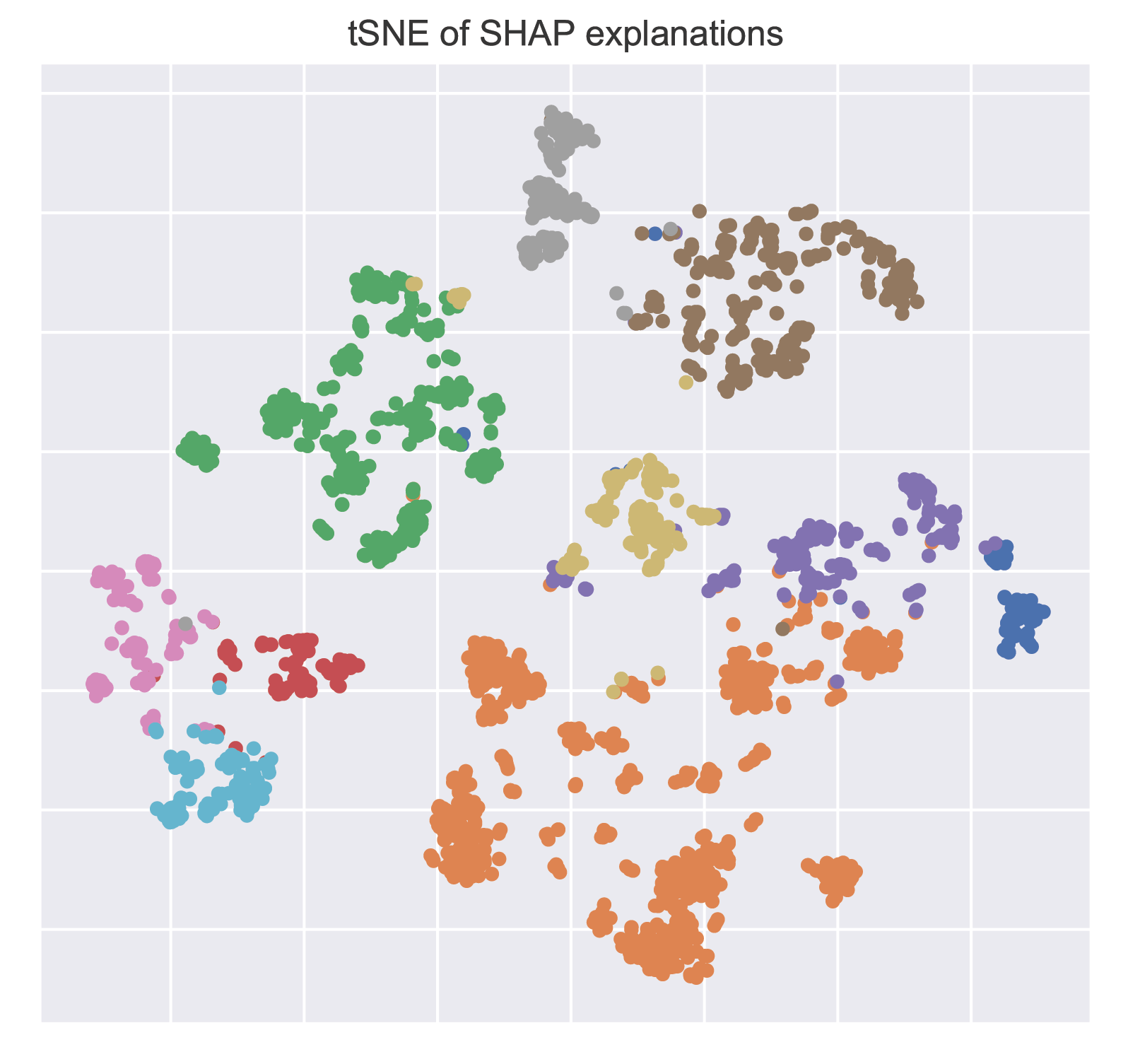}
    \caption{k-means clustering $k=10$}
    \label{fig:clus_rabo}
  \end{subfigure}
  %
  \begin{subfigure}[t]{0.32\textwidth}
    \centering
    \includegraphics[height=\ht1,trim=10 10 10 10,clip]{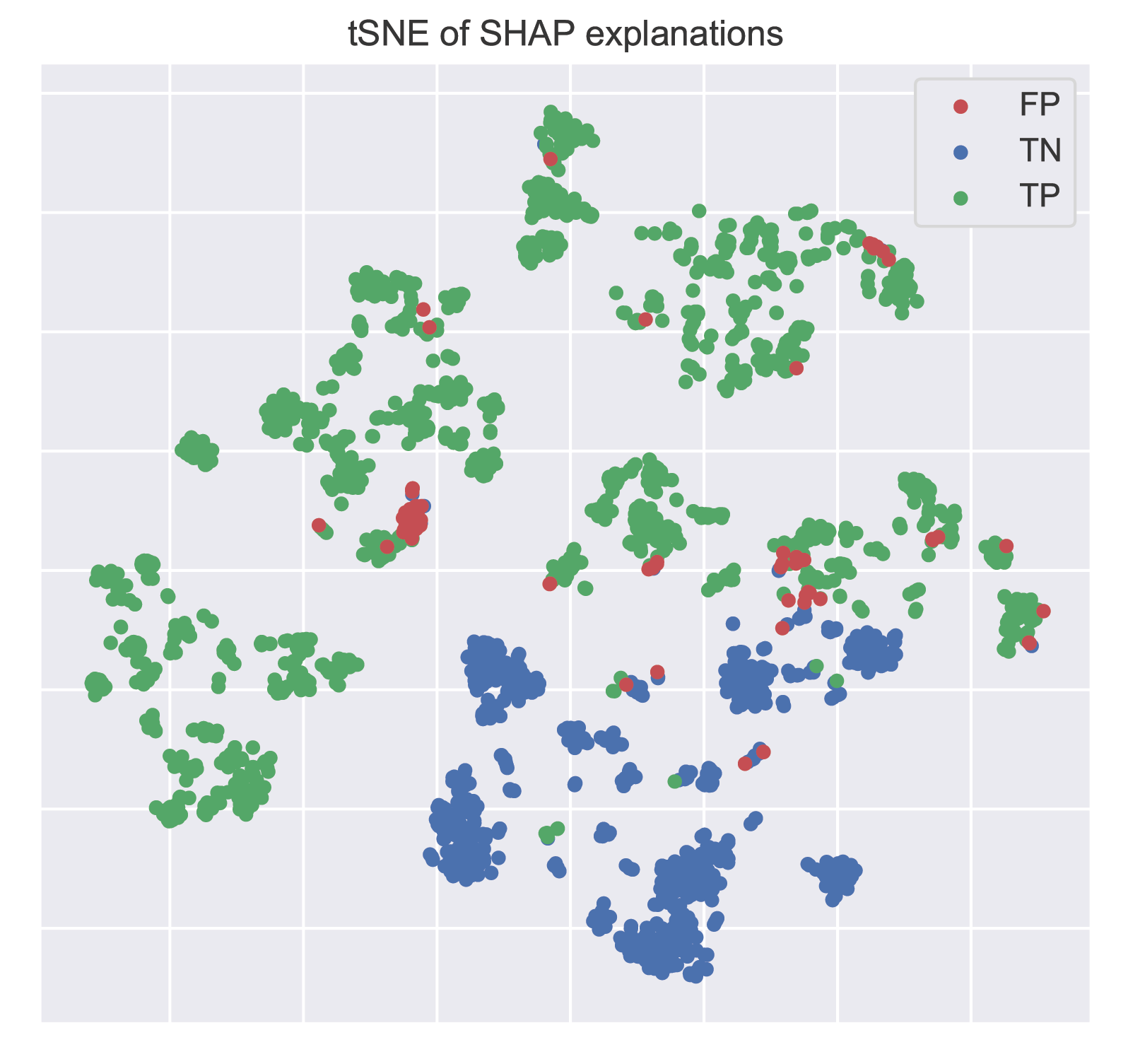}
    \caption{Model's Performance}
    \label{fig:perf_rabo}
  \end{subfigure}
  \caption{t-SNE visualization that groups transactions with similar SHAP explanations. The SHAP explanations explain predictions made by a random forest fraud detection model.}
  \label{fig:clustervisrabo}
\end{figure*}

The idea behind our approach is to retrieve and visualize instances that are similar to the query instance. However, it is unclear which notion of dissimilarity will be most useful for alert processing. Moreover, different combinations of distance functions in the case retrieval and visualization step may be more useful than others.

\subsubsection{Feature Values}
The most straightforward way to define similarity is to consider the feature values of transactions. In this case, instances that with similar feature values are considered similar. Depending on the feature type (e.g. categorical or continuous) different distance functions may be more appropriate than others. In this work, we assume that all feature values are properly normalized and that Euclidean distance is meaningful. However, we encourage readers to use a different distance function if appropriate.

\subsubsection{Feature Contributions}
A potential disadvantage of a plain feature value-based distance function is that the machine learning model is not taken into account at all. Instances that seem similar with regard to feature values, may have been treated very differently by the model. For example, consider a fraud detection decision tree with at its root node the decision rule \texttt{amount $>$ \$10,000}. Two transactions that are exactly the same with regard to all feature values except for \textit{amount} are in different branches of the decision tree. Hence, the transactions may seem very similar in the data, but the decision-making process of the model could be completely different for each of the transactions. Judging the trustworthiness of a new prediction based on instances that were treated differently by the model does not seem intuitive. Instead, it might be more informative to take into account the model's arguments.


A state-of-the-art approach for explaining single predictions are Shapley Additive Explanations (SHAP) \cite{Lundberg2017, Strumbelj2014}. SHAP is based on concepts from cooperative game theory and explains how each feature value of an instance contributed to the model's confidence score. In order to take into account the model's arguments for its predictions, we can consider similarity in SHAP explanations. In a SHAP value-based distance function, instances whose feature values contributed similarly to the model's confidence score are considered similar. 

Interestingly, we find that distances in SHAP value space behave very well. First of all, SHAP explanations can be clustered well using relatively few clusters (see Figure~\ref{fig:clus_rabo}). Moreover, it is possible to identify subsets of transactions for which the model performs worse than for others (see Figure~\ref{fig:perf_rabo}). This indicates that SHAP value similarity can be meaningful for alert processing.

\subsubsection{Globally Weighted Feature Values}
A potential disadvantage of a distance function based solely on SHAP values, is that instances with a similar SHAP explanation can still have different feature values. We can combine SHAP explanations and feature values in a single distance function by weighting features values by the model's feature importance. Feature importance can be defined either globally (i.e. across all instances) or locally (i.e. per instance).

When considering a globally weighted distance function, instances with similar feature values on features that are considered \textit{globally} important by the model (i.e. across all instances in the training data) are considered similar. Global SHAP feature importances can be computed as follows \citep{Lundberg2018}:
\begin{equation}
    \bar{|\Phi|} = \frac{1}{N}\sum_{i=1}^N |\Phi_i|
\end{equation}
where $N$ is the total number of instances in the data set and $\Phi_i$ the SHAP value vector of instance $i$.

\subsubsection{Locally Weighted Feature Values}
Local SHAP importances can be very different from global SHAP importances. For example, a particular feature may be relatively important on average, but not contribute at all to the prediction of a particular instance. Therefore, the utility of feature importance may depend on whether importance is defined globally or locally. When locally weighted feature value distance function is used, instances with similar feature values on features that are considered \textit{locally} important by the model (i.e. for the query instance) are considered similar. Note that this distance function is similar to the one suggested by \citet{Nugent2005}, except we use SHAP importances rather than local feature contributions similar in spirit to LIME.

\subsubsection{Formalization}

As a basic distance function, we consider the weighted Euclidean distance. Given two input vectors $z_a = (z_{a1}, ..., z_{am})$ and $z_b = (z_{b1}, ..., z_{bm})$, and a weight vector $w = (w_1, ..., w_m)$, the weighted Euclidean distance is defined as follows:
\begin{equation}
    \label{eq:weuclidean}
    d_{ab} = d(w, z_a, z_b) = \sqrt{\sum_{j=1}^m w (z_a - z_b)^2}
\end{equation}
Note that Equation~\ref{eq:weuclidean} is equivalent to the unweighted Euclidean distance when $w$ is an all-ones vector ($\mathbf{1}$). We can describe the four considered distance functions with regard to the input vectors of Equation~\ref{eq:weuclidean} (see Table~\ref{tab:distfuncs}).

\begin{table}[ht!]
\caption{In both the \textit{case retrieval} and \textit{neighborhood visualization} stage, we consider four distance functions that differ with regard to input values of Equation~\ref{eq:weuclidean}. $x_a$ refers to the feature value vector of instance $a$, $\Phi_a$ to the SHAP value vector of instance $a$, and $q$ denotes the query instance.}
\label{tab:distfuncs}
\small
\centering
    \begin{tabular}{lll}
    \toprule
    \textbf{Notation} & \textbf{Description}  & \textbf{Definition} \\ \midrule
    $d_F$   & feature values    &  $d(\mathbf{1}, x_a, x_a)$ \\
    $d_S$   & SHAP values       &  $d(\mathbf{1}, \Phi_a, \Phi_b)$  \\
    $d_G$   & \begin{tabular}[c]{@{}l@{}}feature values weighted by\\ \textbf{global} SHAP importance\end{tabular} & $d(\bar{|\Phi|}, x_a , x_b)$  \\ 
    $d_L$   & \begin{tabular}[c]{@{}l@{}}feature values weighted by \\ \textbf{local} SHAP importance \end{tabular}  & $d(|{\Phi_q}|, x_a , x_b)$  \\ \bottomrule
    \end{tabular}
\end{table}
\noindent In the next section we discuss the performance of our approach when $d_F$, $d_S$, $d_G$, and $d_L$ are used. We will omit $d$ and use the corresponding index letter in the figures for brevity.
\section{Evaluation}
\label{sec:eval}

The goal of our CBR approach is to provide evidence on the trustworthiness of a prediction. Similar to \cite{Jiang2018}, we define \textit{local trustworthiness} as the difference between the Bayes-optimal classifier's confidence and the model's prediction (e.g. fraud or no fraud) for that instance. That is, if the model agrees with the Bayes-optimal classifier, trustworthiness is high, and vice versa. Notably, even the Bayes-optimal classifier can be wrong in some regions due to noise. Hence, trustworthiness should be interpreted as an estimate of the reasonableness of the prediction, given the underlying data distribution. In practice, the Bayes-optimal classifier is not realizable and empirical measurements of trustworthiness are not possible. However, our goal is to provide decision support for domain experts that perform alert processing tasks. Hence, we can bypass the difficulty of measuring trustworthiness by measuring utility for domain experts instead.

\subsection{Simulated User Experiment}
\label{sec:evalsim}
We evaluate the expected utility of our visualization for alert processing in a \textit{simulated user experiment}. In a simulated user experiment, assumptions are made about how users utilize an approach to perform particular tasks. Subsequently, the expected task performance is computed as the task performance that is achieved when applying the assumed strategy. Simulated user experiments have been used to evaluate the coverage, precision and trustworthiness of different explanations by \citet{Ribeiro2016, Ribeiro2018}. We are not aware of simulated user experiments aimed at the evaluation of alert processing performance of domain experts. Consequently, we present a novel experiment setup.

In our simulated user experiments, we simulate a situation in which a machine learning model has been put in production. In this scenario, a model has been trained on historical data and new data is arriving. For each new instance, the model predicts whether it is a positive or a negative. Positives will trigger an alert and are inspected by human analysts, while negatives are not further inspected.

The goal of the experiments is to estimate how well a analyst would be able to process alerts when provided with the neighborhood visualization. To this end, we make a few assumptions about how users interpret the visualization. Based on these assumptions, we estimate how confident user would be about the instance belonging to the positive class. In order to determine the utility of the visualization compared to the model, we evaluate how well the estimated user confidence score as well as the model's confidence score correspond to the ground truth.

\subsubsection{Method}
To simulate the scenario where a model has been put into production, we split our data set in three different parts (see Figure~\ref{fig:split}). We can describe the simulated user experiment further using the following six steps:

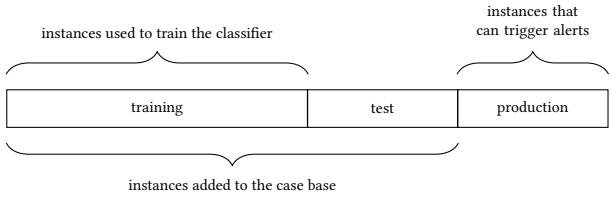
\begin{figure} [ht]
    \scriptsize
    \centering
    \begin{tikzpicture}
        \begin{scope}[start chain=1 going right,node distance=-0.5pt]
            \node(training) [on chain=1, minimum width=4cm, minimum height=0.5cm, draw] {training};
            \node(test) [on chain=1, minimum width=2cm, minimum height=0.5cm, draw] {test};
            \node(production) [on chain=1, minimum width=2cm, minimum height=0.5cm, draw] {production};
        \end{scope}
    
    \draw[decorate, decoration={brace,raise=2mm,amplitude=3mm},draw]
        (training.north west) to 
            node[above=6mm] 
                {instances used to train the classifier} 
        (training.north east);
    
    \draw[decorate, decoration={brace,raise=2mm,amplitude=3mm,mirror},draw]
        (training.south west) to 
            node[below=6mm] 
                {instances added to the case base} 
        (test.south east);
    
    \draw[decorate, decoration={brace,raise=2mm,amplitude=3mm},draw]
        (production.north west) to 
            node[above=6mm, text width=2cm, align=center] 
                {instances that can trigger alerts} 
        (production.north east);
    \end{tikzpicture}
    
    \caption{In the simulated user experiment, the dataset is split into three sets. The \textit{training} and \textit{test} set correspond to data that is available at the time of model inference. The \textit{production} set corresponds to new instances that arrive once the model is in production.}
    \label{fig:split}
\end{figure}

\begin{enumerate}
    \item \textit{Split the dataset.} We first split the dataset into three different sets: the \textit{training data}, the \textit{test data}, and \textit{production data}. The production data represents data that arrives as the model is in production.
    \item \textit{Train classifier.} We train a classifier on the training data.
    \item \textit{Initialize Case Base.} As the training and test set contain instances for which we know the ground truth at the time the model goes into production, we add these instances to the \textit{case base}.
    \item \textit{Initialize Alert Set.} We determine which instances from the production data would result in a positive prediction from our machine learning model. These instances are put in the \textit{alert set}.
    \item \textit{Estimate user's and model's confidence scores.} For each of the instances in the alert set, we estimate the user's confidence of the instance belonging to the positive class as a number between 0 and 1. Additionally, we determine the model's confidence for each instance in the alert set. 
    \item \textit{Evaluate confidence.} Given the ground truth of the instances in the alert set, we compare the mean average precision (MAP) that is achieved using the user's confidence to MAP achieved by the model's confidence.
\end{enumerate}
In order to estimate the user's confidence, we make several assumptions on how our visualization is interpreted. Recall that we are interested in alert processing. We assume that a positive neighbor increases the user's confidence that the instance is a true positive and a negative neighbor decreases the user's confidence. Then, we can estimate the user's confidence, $c_i$ based on the retrieved neighbors using the following equation:
\begin{equation}
    \label{eq:userconfidence}
    c_i = \frac{1}{k} \sum\limits_{j = 1}^k \mathbf{1}\{y_j=1\}
\end{equation}
where $y_j$ is the true class of instance $j$. However, some neighbors may be much more similar to the query instance than others. This is shown to the user in our neighborhood visualization, so the user will likely take this into account. Therefore, we weight each neighbor's class by the inverse distance between the neighbor and the query instance $i$:
\begin{equation}
    \label{eq:userconfidenceweighted}
    c_i^w = \frac{\sum\limits_{j = 1}^k \frac{1}{d_{ij}} \mathbf{1}\{y_j = 1\}}{\sum\limits_{j = 1}^k \frac{1}{d_{ij}}}
\end{equation}
Note that $\frac{1}{d_{ij}}$ is undefined if $d_{ij}$ is equal to zero, i.e. if the neighbor is identical to the instance we are trying to explain. In our experiments, we deal with this by setting $d_{ij}$ to a small number that is at least smaller than the most similar non-identical neighbor. In this way, a large weight to the identical neighbor, but the other neighbors are still taken into account.

 


\subsubsection{Results}
We evaluate our CBR approach on three benchmark classification data sets: \textit{Adult} \citep{adult1997}, \textit{Phoneme} \citep{phoneme1991}, and \textit{Churn} \citep{Olson2017PMLB}. All data sets were retrieved from OpenML \citep{OpenML2013}. Additionally, we evaluate our approach on a real-life \textit{Fraud Detection} data set provided by a major Dutch bank. On each of the data set, we train a random forest classifier using the implementation in scikit-learn \citep{sklearn}.

We evaluate the estimated user confidence scores for each possible combination of distance functions in Table~\ref{tab:distfuncs}. As a baseline, we also add a user confidence score that would be achieved when no distance function is considered in the neighborhood visualization (i.e. Equation~\ref{eq:userconfidence}). These results represent the case in which similarities are not provided to the user at all.

Recall that the number of neighbors $k$ is a user-set parameter. Consequently, the approach is evaluated for different values of $k$, ranging from 1 to 500 neighbors. For each combination of distance functions, we compare the MAP of the model's confidence to the MAP of the estimated user confidence averaged over the different values for $k$. In Figure~\ref{fig:heatmaps}, we summarize the difference in performance as the average over all possible values of $k$.

\subsubsection*{Estimated User Confidence Mostly Performs Better Than Model's Confidence}
\begin{figure*}[ht!]
  \centering
  \begin{subfigure}[t]{0.45\textwidth}
    \includegraphics[width=\textwidth,trim=10 10 10 10,clip]{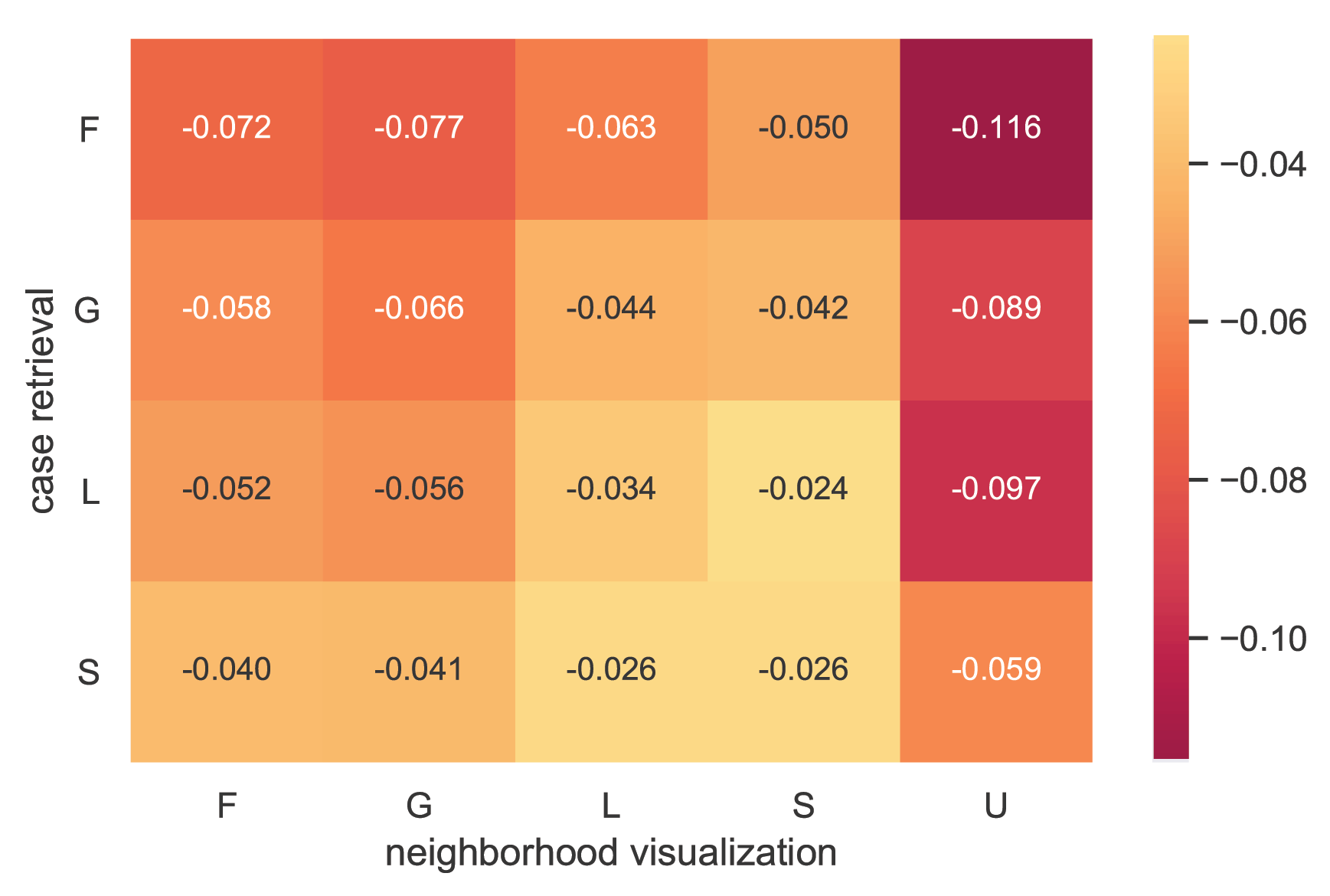}
    \caption{\textit{Adult}}
    \label{fig:heatmap_adult}
  \end{subfigure}
  \hspace{5mm}
  \begin{subfigure}[t]{0.45\textwidth}
    \includegraphics[width=\textwidth,trim=10 10 10 10,clip]{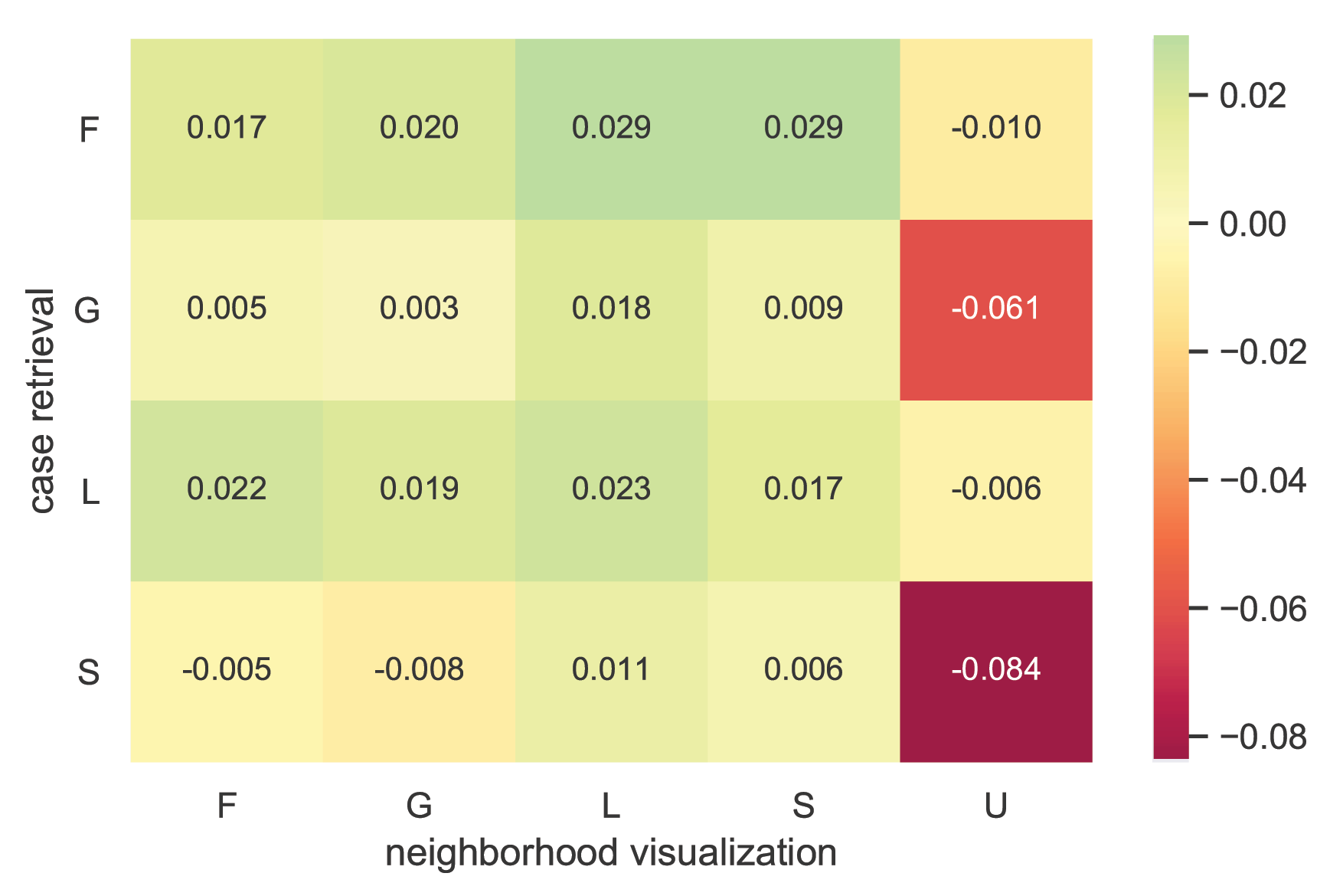}
    \caption{\textit{Churn}}
    \label{fig:heatmap_churn}
  \end{subfigure}
  \begin{subfigure}[t]{0.45\textwidth}
    \centering
    \includegraphics[width=\textwidth,trim=10 10 10 10,clip]{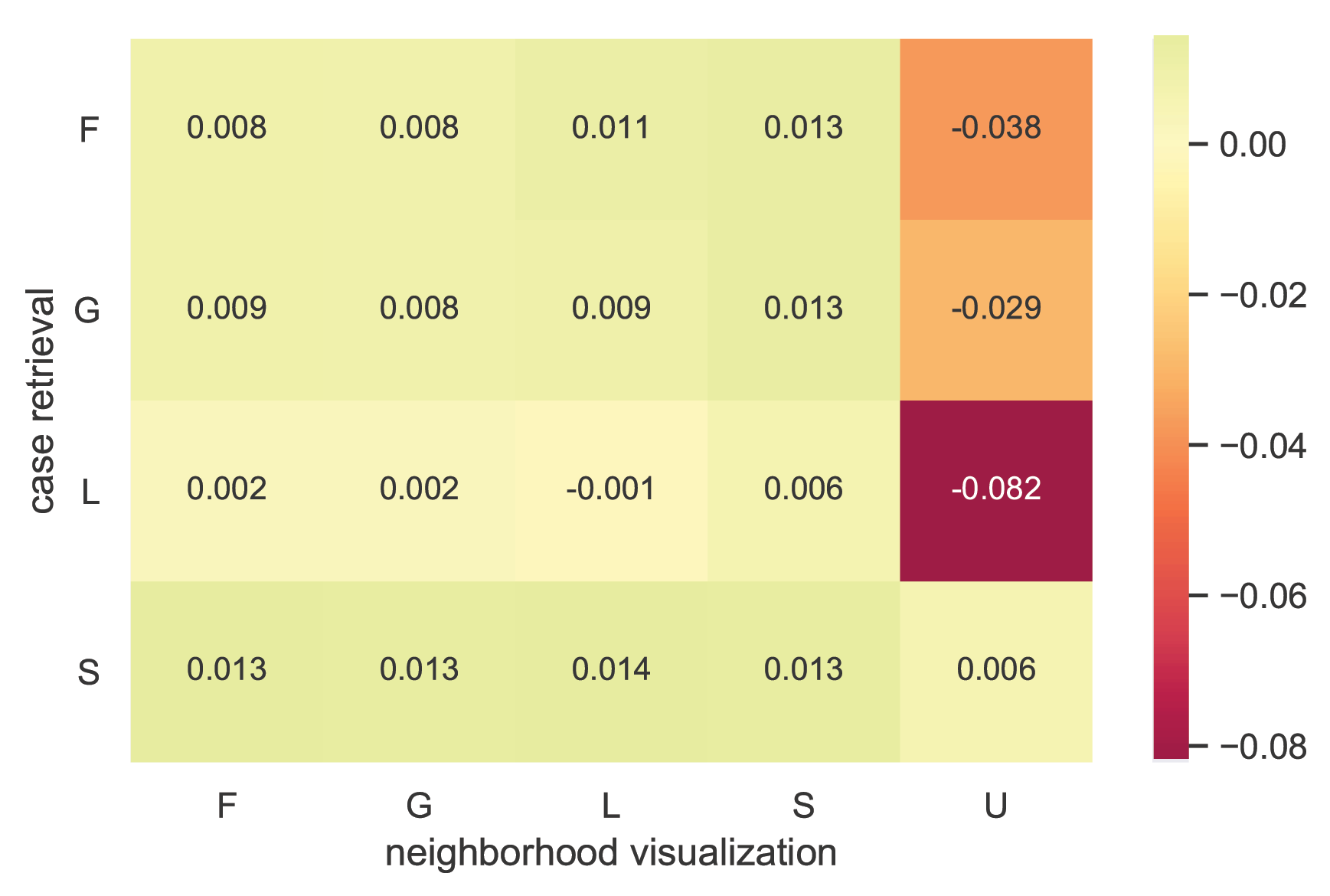}
    \caption{\textit{Phoneme}}
    \label{fig:heatmap_phoneme}
  \end{subfigure}
  \hspace{5mm}
  \begin{subfigure}[t]{0.45\textwidth}
    \centering
    \includegraphics[width=\textwidth,trim=10 10 10 10,clip]{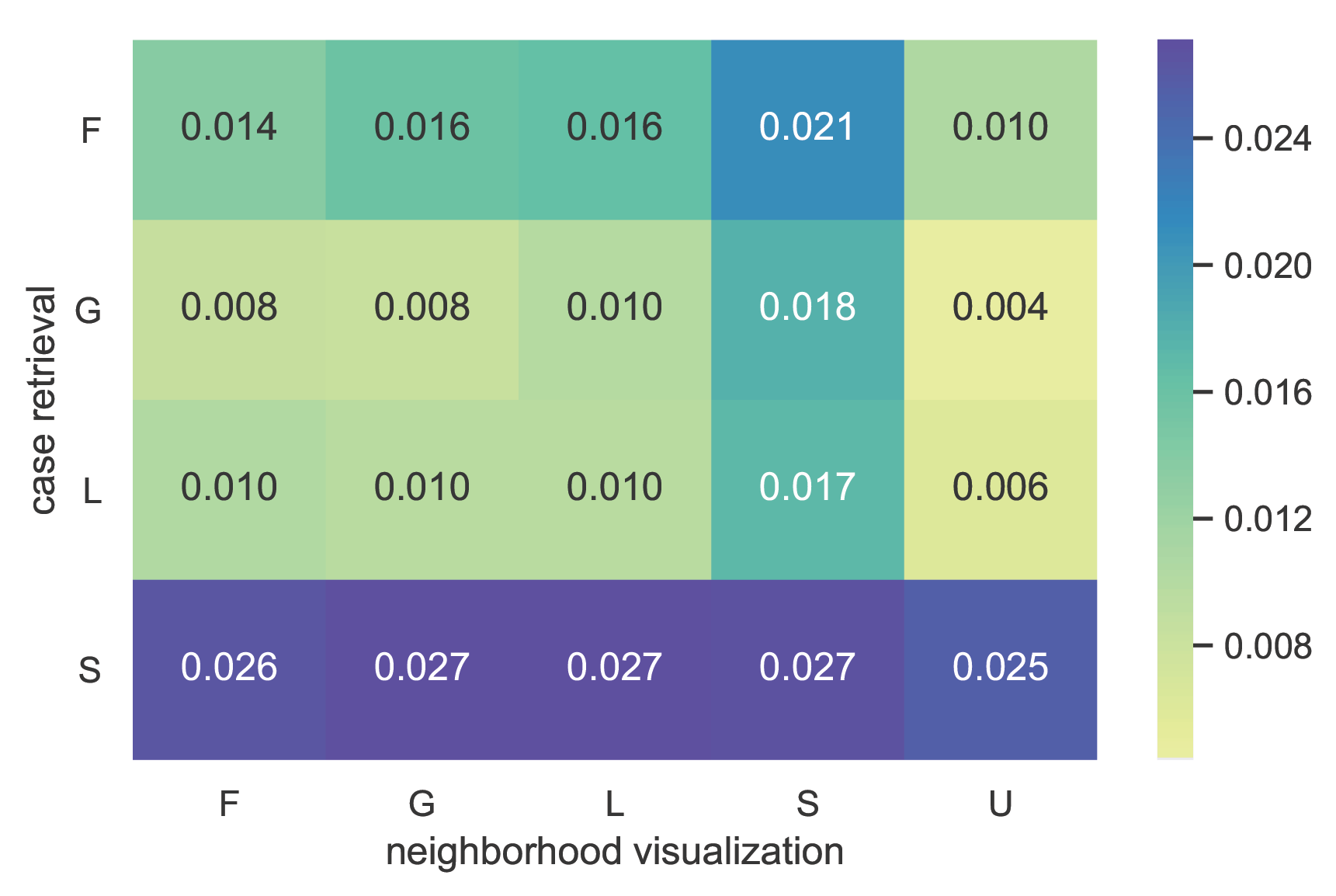}
    \caption{\textit{Fraud Detection}}
    \label{fig:heatmap_fraud}
  \end{subfigure}
  \caption{Improvement or decrease in average MAP of the estimated user confidence score compared to the MAP of the model's confidence score. MAP of the estimated user confidence is averaged over number of retrieved cases $k \in \{1, 2, ..., 500\}$. The difference is shown for all possible combinations of distance functions in the two steps of the approach. $F$, $G$, $L$, and $S$ refer to the distance functions defined in Table~\ref{tab:distfuncs}. $U$ refers to an unweighted estimated user confidence according to Equation~\ref{eq:userconfidence} (i.e. if the user ignores the distances in the neighborhood visualization).}
  \label{fig:heatmaps}
\end{figure*}

For the \textit{Churn}, \textit{Phoneme} and \textit{Fraud Detection} classification tasks, the estimated user confidence mostly results in higher average MAP than the model's confidence, but the achieved performance gain typically differs for different combinations of distance functions (Figure~\ref{fig:heatmaps}). Only for the \textit{Adult} data set, the estimated user confidence results in worse average MAP scores than the model's confidence.

\subsubsection*{Number of Retrieved Neighbors ($k$) Impacts Performance}
\begin{figure}[ht]
  \centering
  \begin{subfigure}[t]{0.4\textwidth}
    \includegraphics[width=\textwidth,trim=10 10 10 10,clip]{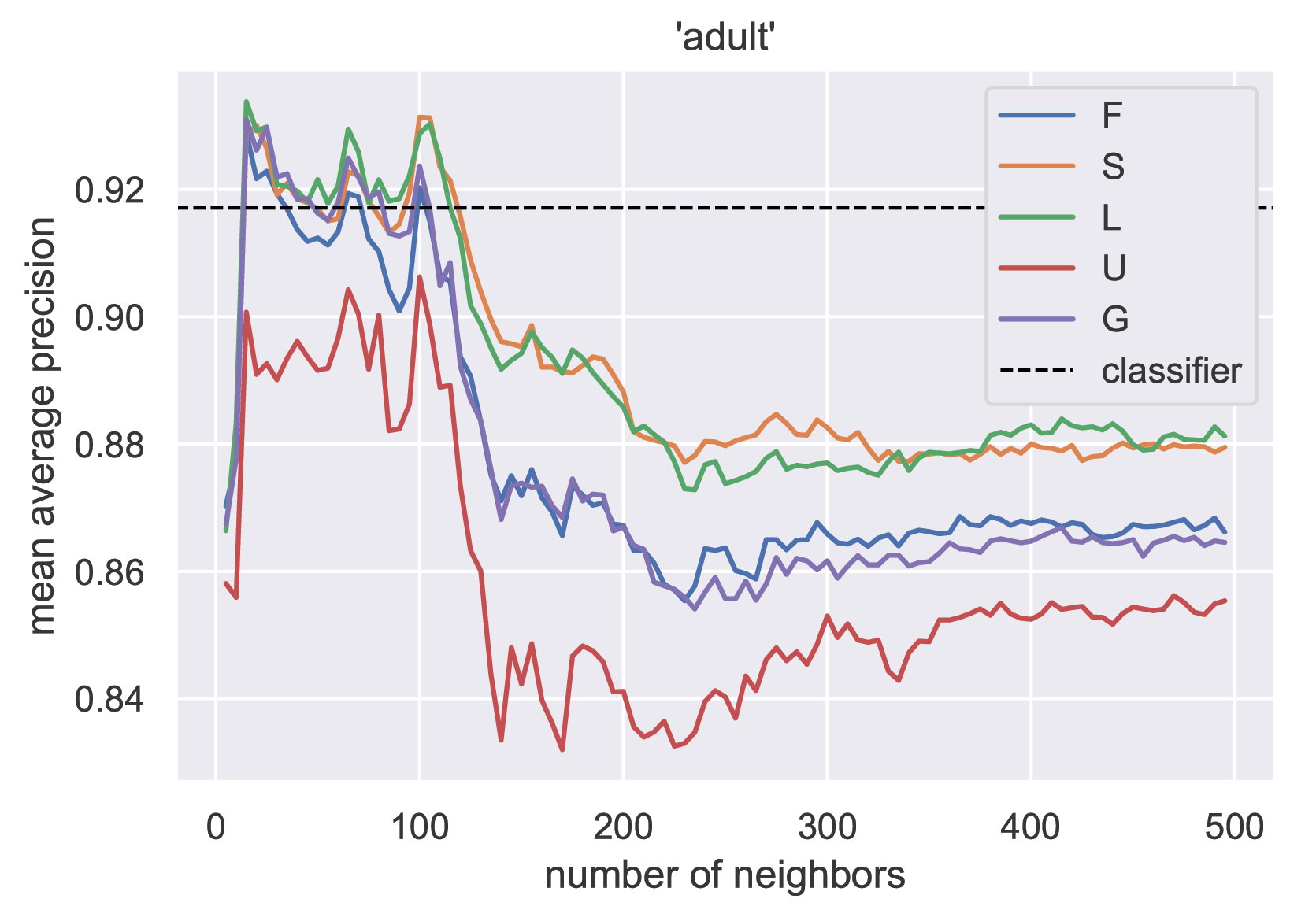}
    \caption{\textit{Adult}}
    \label{fig:cbrperf_adult}
  \end{subfigure}
  \hspace{5mm}
  \begin{subfigure}[t]{0.4\textwidth}
    \includegraphics[width=\textwidth,trim=10 10 10 10,clip]{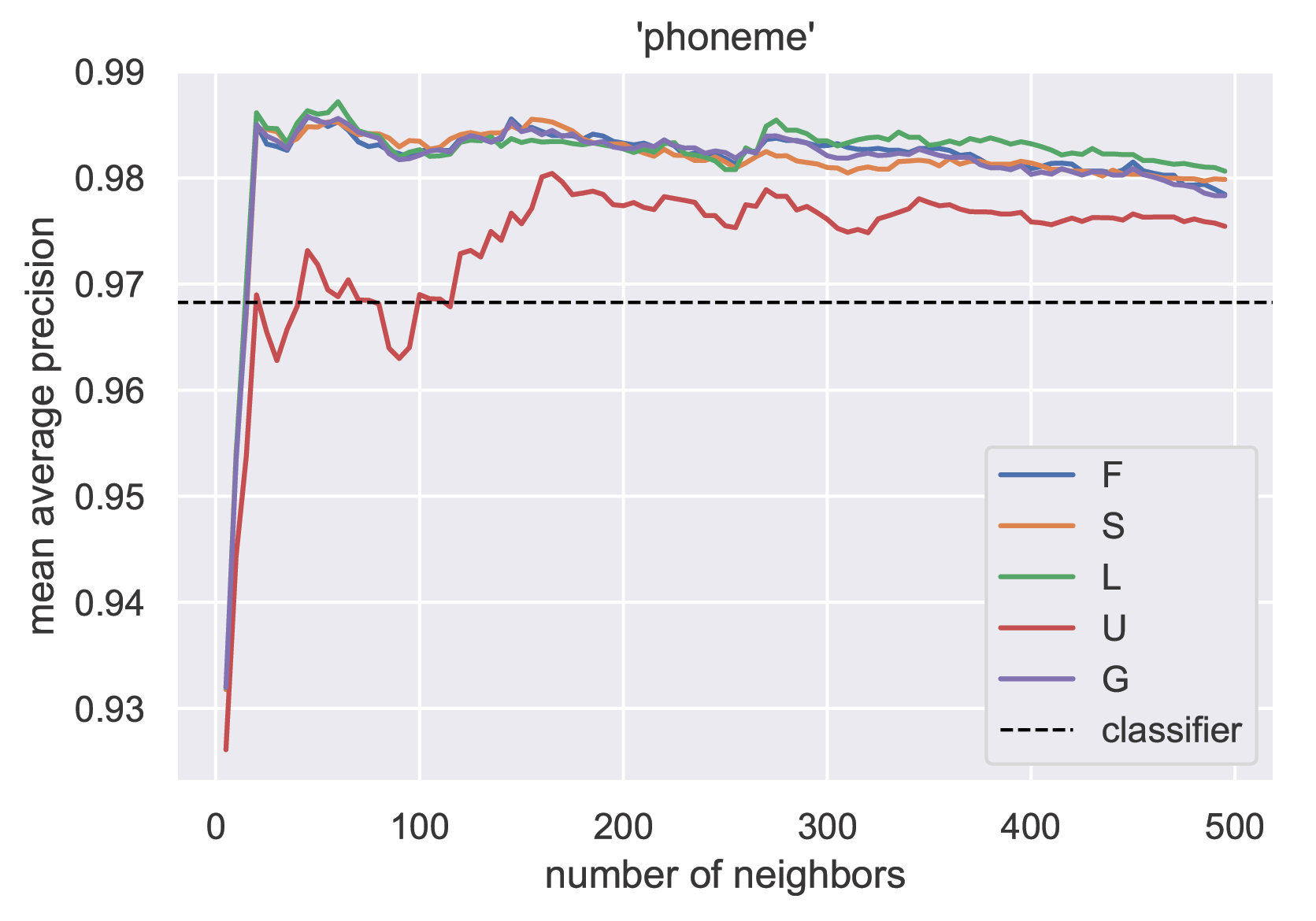}
    \caption{\textit{Phoneme}}
    \label{fig:cbrperf_phoneme}
  \end{subfigure}
  \caption{The performance of different neighborhood visualization functions against the retrieved number of neighbors for the \textit{Adult} and \textit{Phoneme} data set when using $d_S$ for case retrieval.}
  \label{fig:n_neighbors}
\end{figure}

In some data sets, user-set parameter $k$ has a high impact on the performance of the estimated user confidence. In particular, our approach outperforms the classifier in the \textit{Adult} data set only for a very particular range of neighbors (see Figure~\ref{fig:cbrperf_adult}). For the \textit{Phoneme} and \textit{Fraud Detection} data sets, a minimum number of neighbors of approximately 20 is typically required to outperform the model's confidence score (see Figure~\ref{fig:cbrperf_phoneme}). This result suggests that returning only the most similar case to the user, as suggested by \citet{Nugent2005}, may not provide enough evidence to be useful for alert processing. When applied to new problems, simulated user experiments could be performed to decide upon the appropriate range of $k$ that can be selected by a real human user.

\subsubsection*{Unweighted User Confidence Performs Consistently Worse than User Confidence Weighted By Any Distance Function}
For each of the data sets, estimating the user's confidence as the simple average of the true class of the retrieved neighbors consistently results in the worst performance. Recall that unweighted user confidence corresponds to a user who ignores similarity of the retrieved cases. This result shows the importance of communicating the similarity of the retrieved neighbors to the user, as is done in the \textit{neighborhood visualization} step.

\subsubsection*{$d_S$ Mostly Performs Best}
For all data sets apart from the \textit{Churn} data set, using $d_S$ performs best for both case retrieval and neighborhood visualization (see Figure~\ref{fig:heatmaps}). In particular, performing case retrieval using $d_S$ for the \textit{Phoneme} and \textit{Fraud Detection} data sets consistently results in top performance, regardless of the distance function that is used in neighborhood visualization. This indicates that the relevance of the retrieved neighbors is very high. In the \textit{Churn} data set, $d_F$ and $d_L$ perform best for case retrieval and $d_L$ for neighborhood visualization.



\subsection{Usability Test}
\label{sec:evalusa}
To determine the perceived utility of our approach for fraud analysts at the Rabobank, we conduct usability test. 

\subsubsection{Method}
\begin{figure*}[ht]
    \centering
    \includegraphics[width=\linewidth]{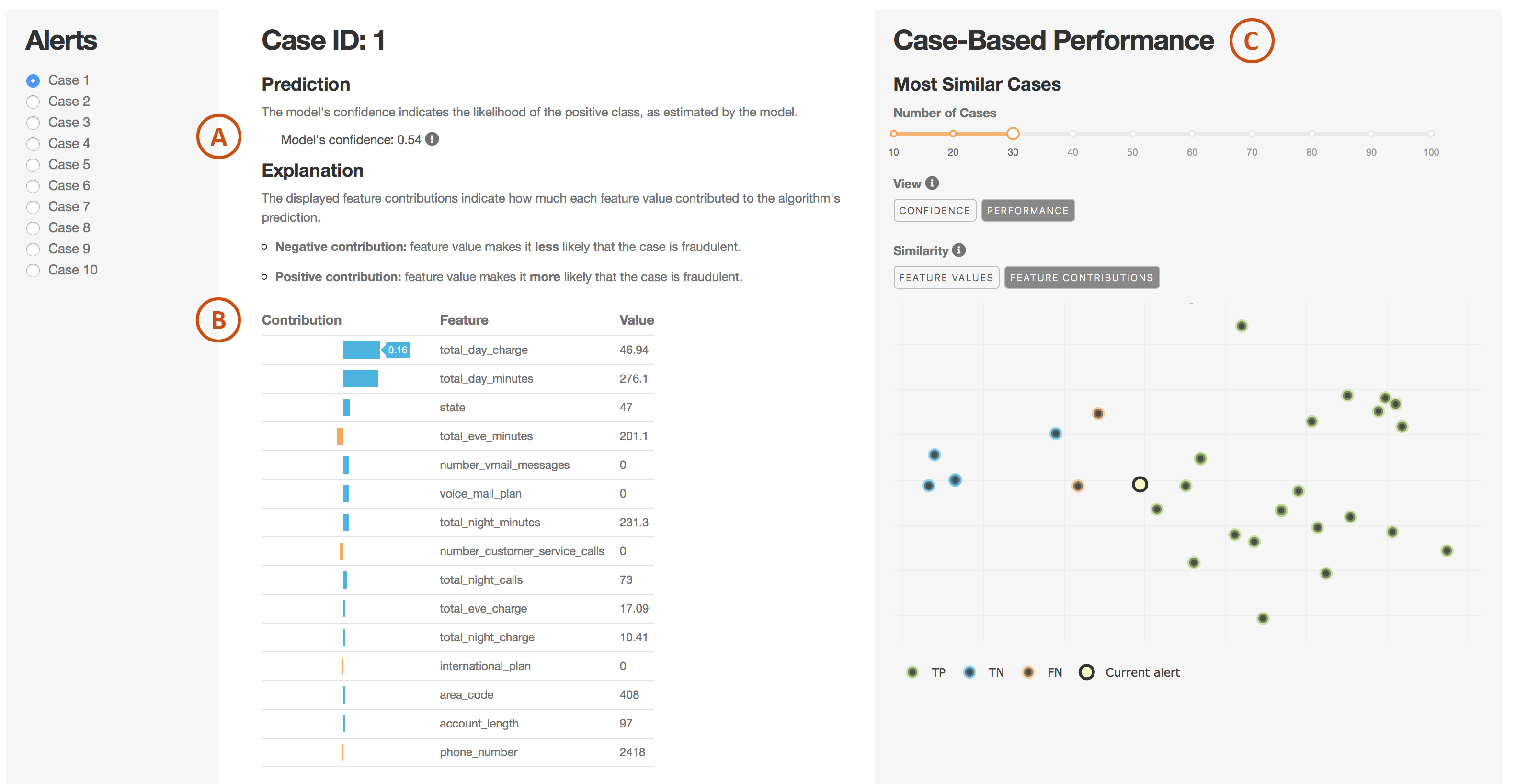}
    \caption{CBR dashboard when applied to predictions of a random forest model trained on the \textit{Churn} dataset. (A) The model's confidence for the selected alert, (B) A bar chart showing the SHAP values of the selected alert, (C) The CBR neighborhood visualization. The number of neighbors $k$ can be chosen in the slider.} 
    \label{fig:dashboard}
\end{figure*}

The CBR approach is implemented in a Python-based dashboard, which displays the model's confidence, SHAP explanation, and neighborhood visualization of a selected alert (see Figure~\ref{fig:dashboard}). The evaluation is performed in individual sessions with fraud analysts, using a think-out loud protocol. After the usability test, the dashboard is evaluated on \textit{perceived usefulness} and \textit{perceived ease of use} by means of a short survey introduced by 
\citet{Davis1989}.


\subsubsection{Results}
Four fraud analysts participated in the evaluation. The average perceived usefulness was $5.64$ on a 7-point Likert scale, with a standard deviation of $1.2$. The average perceived utility was $5.96$ out of 7, with a standard deviation of $0.9$.

From the verbal protocols, it became clear that the neighborhood visualization materializes the fraud analysts' intuitions on the trustworthiness of fraud detection rules. As such, we expect the system to be particularly relevant for performing deeper analyses of cases for which a fraud analyst has not yet developed a strong intuition. As fraud detection models are constantly retrained, explanations for machine generated alerts are expected to differ over time, which makes our approach particularly relevant in that scenario.

\section{Conclusions}
\label{sec:concl}
Recent explanation methods have been proposed as a means to assess trust in a model's prediction. However, there is a lack of empirical evidence that illustrates the utility of explanations for alert processing tasks. In particular, understanding why the model made a certain prediction may not be enough to assess the correctness of the prediction. Hence, rather than explaining a prediction, our goal is to provide evidence on the reasonableness of the prediction given the underlying data. In this paper, we have introduced a novel CBR approach that can be used to assess the \textit{trustworthiness} of a prediction.

In our simulated user experiments, we have shown that the two-stage CBR approach can be useful for processing alerts. According to our intuitions, our results suggest that a distance function based on similarity in SHAP values is more useful than distances based on feature value similarity. Moreover, the results of a usability test with fraud analysts at a major Dutch bank indicate that our approach is perceived useful as well as easy to use.


\subsection{Future Work}
In the present paper, we have evaluated our approach on four different classification tasks with some varying results. Not all of these results are already well understood. In particular, future work could consider a more extensive analysis on why particular distance functions work well for some data sets and not as good for others. 

Additionally, future work could consider extensions of the neighborhood visualization. In particular, adding counterfactual instances is expected to provide more insights in the decision boundary of the model.

As SHAP values can be expensive to compute, future work could focus on optimizing the case base, by means of e.g. prototype selection or sampling approaches. An important aspect of these approaches is how they may be misleading for users. In particular, future work could study how over- and undersampling approaches affect the decision-making process of users in scenarios with highly imbalanced data.

One of the findings presented in this work is that SHAP explanations are remarkably clusterable. An interesting direction of future work that leverages this observation are \textit{prototypical explanations}, which could be used to provide a global explanation of a black-box model in a model-agnostic fashion.





\bibliographystyle{ACM-Reference-Format}
\bibliography{references}

\appendix

\end{document}